\documentclass[11pt]{article}

\usepackage[preprint]{acl}

\usepackage{times}
\usepackage{latexsym}

\usepackage[T1]{fontenc}

\usepackage[utf8]{inputenc}

\usepackage{microtype}

\usepackage{inconsolata}

\usepackage{graphicx}

\usepackage{booktabs}
\usepackage{caption}
\usepackage{multirow}
\usepackage{subcaption}

\usepackage{url}
\usepackage{hyperref}

\usepackage{amsmath}
\usepackage{mathtools}
\usepackage{amssymb}
\usepackage{amsthm}
\usepackage{bbding}

\title{Learning New Facts with QLoRA: An Acquisition-Retention Frontier}

\author{
 \textbf{Estelle Zheng\textsuperscript{1,2}},
 \textbf{Sébastien Warichet\textsuperscript{2}},
 \textbf{Emmanuel Helbert\textsuperscript{2}},
 \textbf{Christophe Cerisara\textsuperscript{1}}
\\
 \textsuperscript{1}LORIA, CNRS, France \\
 \textsuperscript{2}Alcatel-Lucent Enterprise, France \\
    \texttt{ \{estelle.zheng, cerisara\}@loria.fr} \\
    \texttt{ \{sebastien.warichet, emmanuel.helbert\}@al-enterprise.com}
}

\begin{document}
\maketitle
\begin{abstract}


Parameter-efficient fine-tuning is often assumed to preserve pretrained
capabilities because it updates only a small number of parameters. We show that
this assumption depends strongly on adapter capacity. We study factual
acquisition in a controlled OpenStreetMap-derived benchmark where Qwen3-4B must
acquire anonymized geographic associations while retaining unrelated capabilities.
Comparing full fine-tuning (FFT) with quantized low-rank adaptation (QLoRA) at
ranks 8, 16, 32, and 64, we find that rank
induces a clear acquisition--retention frontier. Low-rank QLoRA preserves
out-of-domain (OOD) performance but acquires fewer facts, whereas higher ranks improve
same-fact paraphrase generalization at an increasing cost in performance on
unrelated benchmarks.
FFT behaves as a conservative baseline: it retains general
capabilities well, but does not reach the highest factual-acquisition regime.
Distributional, weight-space, and spectral diagnostics mirror this behavioral
trade-off, with higher-rank QLoRA moving farther from the pretrained model. A
separate math adaptation experiment shows a weaker frontier, suggesting that the
effect is most pronounced when adaptation must install new factual associations
rather than reinforce skills already supported by pretraining.\footnote{Code and data
are available at
\url{https://github.com/zhngstl/new_facts_forgetting}.}
\end{abstract}

\section{Introduction}
\label{sec:introduction}

Pretrained language models (PLMs) are often fine-tuned for new
domains, task-specific skills, or factual knowledge. Full fine-tuning (FFT)
updates all model parameters and can be effective, but it is costly and may
degrade performance outside the adaptation distribution. Parameter-efficient fine-tuning (PEFT)
methods such as low-rank adaptation (LoRA) and its quantized variant QLoRA
reduce this cost by freezing pretrained
weights and learning low-rank updates~\citep{hu2022lora,dettmers2023qlora}.
This restriction is often assumed to improve retention of previous capabilities, but it may also limit
what the model can acquire.

Recent work shows that the relationship between LoRA and FFT is not simply one
of efficiency. \citet{biderman2024lora} find that LoRA can preserve more
out-of-domain (OOD) behavior partly because it learns less from the target
distribution. \citet{shuttleworth2024lora} show that LoRA and FFT can reach
similar accuracy from different regions of weight space.
Other comparisons study broad
adaptation settings such as coding~\citep{mannisto-etal-2025-comparative},
mathematics~\citep{biderman2024lora}, question answering
\citep{sun2023comparativestudyfullparameterlorabased}, or instruction tuning
\citep{xin-etal-2024-beyond}. These settings mix several gains:
format adaptation, skill reinforcement, domain shift, or new information. 
We focus next on factual knowledge acquisition.
In this context, retention is ambiguous unless
acquisition is measured at the same time: a low-capacity adapter may appear
safer simply because it has not strongly incorporated the target facts.




Our setting relates to factual knowledge editing, which modifies specific
associations while preserving unrelated behavior
\citep{meng2022locating,mitchell2022fast,meng2023mass,yang2025finetuning}, and
continual learning, which studies the stability--plasticity trade-off under
sequential updates~\citep{jang2022towards,shi-etal-2025-continual}. However,
model-editing benchmarks often focus on localized modifications of previously
known facts, sometimes replacing existing associations, while continual-learning
approaches commonly evaluate sequences of tasks or
updates, including PEFT-based methods that regularize, initialize, or merge
adapter subspaces~\citep{lu-etal-2025-controlled,qiao2026merge}. Our goal is
different: we study a single controlled batch-adaptation stage in which models
acquire novel factual associations using standard FFT and QLoRA. We measure how
adaptation capacity affects acquisition,
paraphrase generalization, and retention of general LLM capabilities.

We introduce a factual acquisition benchmark derived from OpenStreetMap (OSM) in which
models are trained to acquire anonymized geographic associations.
The anonymized entities reduce direct reliance on pretrained world
knowledge, while the OSM structure preserves realistic relational dependencies.
We compare FFT with QLoRA ranks $r \in \{8,16,32,64\}$ and evaluate supervised
fact memorization, same-fact paraphrase generalization, and OOD
retention. We further connect behavioral performance to model-drift diagnostics,
including KL divergence from the base model, RMS-normalized dense update norms,
and SVD-based spectral changes. A standard-LoRA rank sweep on Qwen3-1.7B
separately tests whether the rank trend persists without quantization.

Our results show that LoRA rank induces a clear acquisition--retention
frontier. Low-rank LoRA preserves OOD performance but acquires fewer
facts,
whereas higher-rank LoRA improves factual acquisition at the cost of larger OOD
degradation.
FFT behaves as a baseline: it retains general
capabilities relatively well, but does not reach the highest factual-acquisition
regime observed with higher-rank LoRA. The same trend appears in model-drift
diagnostics, where higher-acquisition LoRA runs move farther away from the
pretrained model.

This paper makes three contributions: (i) we introduce a controlled
OpenStreetMap-derived benchmark for factual acquisition, using anonymized
entities to reduce direct reliance on pretrained world knowledge; 
(ii) we show
that QLoRA rank controls an acquisition--retention frontier for new factual
associations; 
and (iii) we connect this behavioral trade-off to model-drift
diagnostics, showing that stronger factual acquisition is associated with
larger distributional and weight-space shifts from the pretrained model.

\section{Methodology}
\label{sec:methodology}



Standard fine-tuning datasets often evaluate broad task adaptation rather than
the acquisition of genuinely new facts. Benchmarks commonly used to evaluate
knowledge editing, such as
ZsRE~\cite{levy-etal-2017-zero}, CounterFact~\cite{meng2022locating},
MQuAKE~\cite{zhong-etal-2023-mquake}, and
RippleEdits~\cite{cohen-etal-2024-evaluating} typically evaluate localized
updates to known facts, including counterfactual or outdated associations. They
are complementary to our goal of studying standard adaptation on a batch of
novel anonymized associations. A fully synthetic benchmark could also provide
novel facts, but its topology, relation frequencies, and cross-relation
dependencies would have to be chosen by the researcher. We instead use
OpenStreetMap (OSM) because it supplies a naturally occurring, internally
coherent graph whose structure was generated independently of our experimental
hypotheses. Anonymization then reduces reliance on pretrained lexical knowledge
while retaining this non-uniform relational structure.

\subsection{OSM Factual Acquisition Dataset}
\label{sec:short-osm}

We derive atomic facts from 14 city-level OSM extracts, linking entities
(POIs, roads, and cities) to five relation types: POI category, containing city,
nearest road, nearest POI, and road-length bucket. The training split contains
1,938 instruction-style question-answer examples covering direct queries,
paraphrases, locality-preservation probes, spatial-compositional questions, and
inverse city-signature examples. Evaluation uses 900 held-out examples derived
from the same facts but expressed with disjoint surface templates, so performance
measures the acquisition of 
factual associations and their
generalization across surface forms rather than prompt memorization. Because some
relations have distinct answer types, the benchmark does not by itself establish
that models learn abstract relation semantics.

To reduce contamination from pretrained world knowledge, we restrict source
cities to small cities and replace all entity names with synthetic identifiers
(e.g., \texttt{C-TRAIN-001}, \texttt{POI-TRAIN-000001}). Dataset details are in
Appendix~\ref{sec:dataset}, and example prompts are in
Appendix~\ref{app:dataset-examples}.


\subsection{Base-model prior knowledge diagnostic}

Before fine-tuning, we test whether the base model can already solve the task
from prior knowledge or answer-type biases. We evaluate both anonymized and
non-anonymized versions of the data as a question-answering task, where the
model is prompted to generate the gold answer. We report exact-match (EM)
generation accuracy and a teacher-forced gold-vs-distractor preference score.
For each example, we sample five distractors from other gold answers in the
same split, matching both answer type and relation whenever possible. The model prefers the gold
answer when its average per-token log-probability exceeds that of the
distractor. We report the percentage of gold-preferred pairs and the mean
log-probability margin ($\Delta$lp). More details are in Appendix~\ref{app:metrics}.
\begin{table}[t]
\centering
\small
\begin{tabular}{llccc}
\toprule
Split & Names & EM & Pref. (\%) & $\Delta$lp \\
\midrule
Train facts     & Non-anon. & 8.10  & 71.23 & 3.87 \\
Train facts     & Anon.     & 2.43  & 58.46 & 0.26 \\
Paraph. eval    & Non-anon. & 15.67 & 68.98 & 4.02 \\
Paraph. eval    & Anon.     & 9.89  & 59.33 & 0.84 \\
\bottomrule
\end{tabular}
\caption{
\textbf{Base-model prior diagnostic.}
Anonymization reduces EM accuracy, gold-answer preference, and
log-probability margins, suggesting that real names activate relevant pretrained
information. The higher paraphrase EM
partly reflects its larger share of constrained-response questions; see
Appendix~\ref{app:response-formats}.
}
\label{tab:base-prior}
\end{table}

Table~\ref{tab:base-prior} shows that real entity names provide useful semantic
cues, while anonymization sharply reduces exact match and answer-likelihood
margins. Preference scores remain slightly above chance, indicating weak
structural or answer-type biases, but the base model cannot solve the
anonymized task directly. The higher EM on the paraphrase split is partly a
response-format effect. The paraphrase split contains roughly twice the proportion of
yes/no questions, increasing its approximate chance EM from 6.68\% to 10.36\%.
Appendix~\ref{app:response-formats} gives the complete counts and ratios.

\section{Experimental Setup}
\label{sec:experimental-setup}

\subsection{Models and adaptation methods}
We use Qwen3-4B~\citep{yang2025qwen3technicalreport} as the base model and
compare full fine-tuning (FFT) with QLoRA adapters of rank
$r \in \{8,16,32,64\}$. Each training example consists of a question and its
gold answer, with the autoregressive loss applied only to answer tokens. All
runs are repeated over five random seeds. Hyperparameters are reported in
Appendix~\ref{app:hyperparameters}.

To test whether the within-adapter rank trend persists without quantization, we
additionally run standard LoRA on Qwen3-1.7B~\citep{yang2025qwen3technicalreport}
at ranks $r\in\{8,16,32\}$, using the same OSM task and OOD evaluation suite.
This reduced control changes model scale; within its rank sweep.


\subsection{Evaluation axes}
We evaluate each adapted model along three axes.

\paragraph{Factual acquisition} 
We report EM accuracy on two OSM splits. Training accuracy measures
recovery of the supervised facts, while paraphrase accuracy measures same-fact
generalization under held-out templates disjoint from training. Because the
paraphrase set is derived from training facts, it does not test unseen OSM
knowledge; rather, it tests whether the learned association
is robust to phrasing variation.


\paragraph{OOD retention} 
We use LM Evaluation Harness~\citep{eval-harness} on five benchmarks:
HumanEval~\citep{chen2021evaluating}, IFEval~\citep{zhou2023instruction},
TruthfulQA~\citep{lin-etal-2022-truthfulqa},
MMLU-Redux-2.0~\citep{gema-etal-2025-done}, and
BBH~\citep{suzgun-etal-2023-challenging}. These cover code generation,
instruction following, truthfulness, general knowledge, and reasoning. We define
forgetting as the drop in average OOD score relative to the base model:
\[
\Delta_{\mathrm{OOD}}
=
\mathrm{OOD}_{\mathrm{base}}
-
\mathrm{OOD}_{\mathrm{adapted}}.
\]


\paragraph{Model-drift diagnostics}
Behavioral accuracy alone does not reveal how acquisition is achieved: two
models can reach similar OSM accuracy while differing substantially in how far
they move from the pretrained model, with different implications for retention.
Following prior work on LoRA retention and LoRA--FFT weight-space differences
\citep{biderman2024lora,shuttleworth2024lora}, we therefore measure drift using
KL divergence from the base model~\citep{shenfeld2026rls}, teacher-forced
negative log-likelihood on gold OSM answers, RMS-normalized dense weight drift,
and SVD-based intruder dimensions~\citep{pmlr-v9-glorot10a,shuttleworth2024lora}.
For comparability, FFT and QLoRA are analyzed in the same dense update space:
$W_{\mathrm{ft}} - W_0$ for FFT and
$\Delta W = \frac{\alpha}{r}BA$ for QLoRA. RMS normalization controls for
differences in module size. Full metric definitions are given in
Appendix~\ref{app:metrics}.

\begin{figure}[htbp]
    \centering
    \includegraphics[width=0.98\linewidth]{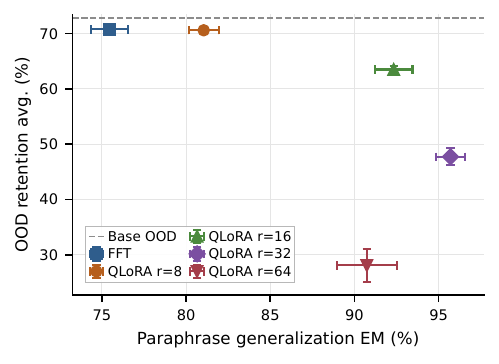}
    \caption{
    \textbf{OSM paraphrase accuracy against average OOD performance.} 
    Points show final-checkpoint means and error bars show standard deviations
    over five seeds.
    Higher-rank QLoRA reaches stronger
    acquisition but lower retention, while FFT and rank 8 remain closer to the
    pretrained model.
    }
    \label{fig:acquisition-retention}
\end{figure}

\section{Results}
\label{sec:results}

\begin{figure*}
    \centering
    \includegraphics[width=0.98\linewidth]{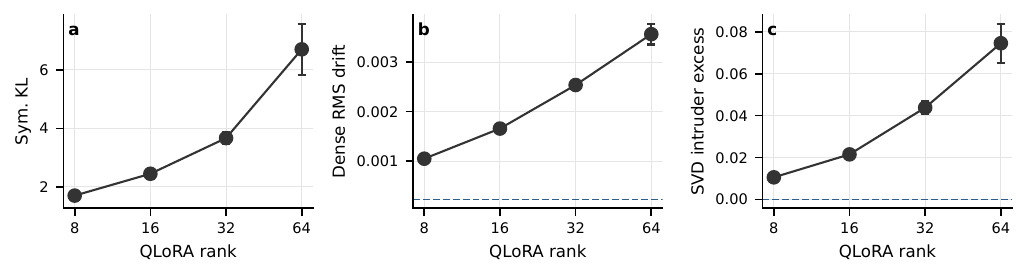}
    \caption{
    \textbf{Model-drift diagnostics for different QLoRA ranks.} (a) Higher-rank QLoRA adapters show larger KL divergence from 
    the pretrained model, (b) larger effective weight updates, and (c) larger spectral shifts 
    under the SVD intruder diagnostic.
    Dashed lines show FFT for comparison. Points show means and error bars show
    std over five seeds.}
    \label{fig:drift-diagnostics}
\end{figure*}

\subsection{QLoRA rank controls the acquisition--retention trade-off}


Figure~\ref{fig:acquisition-retention} shows that QLoRA rank acts as a
plasticity control. Low rank keeps the model close to the pretrained solution
and therefore preserves OOD behavior, but this retention coincides with weaker
same-fact generalization. Increasing rank allows the model to install the OSM
associations more reliably, but moves it onto a lower-retention part of the
frontier. Rank 64 occupies a high-plasticity, low-retention regime: factual
accuracy remains high, but unrelated capabilities collapse. Thus, QLoRA is not
uniformly safer than FFT; its behavior depends on where rank places the model
on the acquisition--retention frontier. The per-benchmark results in
Appendix~\ref{app:ood-breakdown} show that degradation is broad on HumanEval,
IFEval, MMLU-Redux, and BBH, while TruthfulQA remains comparatively stable.

\paragraph{Standard-LoRA control.}
The unquantized Qwen3-1.7B control shows the same qualitative monotonic trade-off:
paraphrase EM rises from 76\% at $r=8$ to 79\% at $r=16$ and 86\% at $r=32$,
while average OOD performance falls from 57.0\% to 52.0\% and 40.2\%,
respectively. This suggests that quantization is not required for the qualitative
rank trend, although this reduced control changes model scale.

\subsection{Higher acquisition requires greater adaptation capacity}
Endpoint comparisons can conflate adaptation method with achieved task
performance: a method may appear to retain more simply because it has
acquired fewer target facts. We therefore compare, for each method and seed, the
evaluated checkpoint closest to three target paraphrase accuracies in
Table~\ref{tab:matched-acquisition}.

FFT and QLoRA $r=8$ retain OOD performance well but do not reach the highest
paraphrase accuracy. Higher-rank QLoRA configurations achieve stronger
paraphrase performance only with larger OOD losses.
This suggests that the apparent robustness of low-rank adaptation to forgetting is actually partly due to limited
plasticity.


\begin{table}[htbp]
\centering
\small
\begin{tabular}{lrrrrrr}
\toprule
& \multicolumn{2}{c}{Target 75} 
& \multicolumn{2}{c}{Target 85}
& \multicolumn{2}{c}{Target 95} \\
\cmidrule(lr){2-3}
\cmidrule(lr){4-5}
\cmidrule(lr){6-7}
Method & Para. & Ret. & Para. & Ret. & Para. & Ret. \\
\midrule
FFT        & 74.9 & 97.6 & 76.1 & 96.9 & 76.1 & 96.9 \\
QL r=8  & 75.7 & 97.9 & 79.4 & 97.5 & 79.4 & 97.5 \\
QL r=16 & 79.3 & 94.3 & 85.8 & 91.1 & 93.3 & 89.9 \\
QL r=32 & 85.2 & 79.8 & 85.4 & 83.0 & 94.6 & 73.3 \\
QL r=64 & 81.0 & 36.7 & 83.1 & 34.3 & 90.4 & 38.8 \\
\bottomrule
\end{tabular}
\caption{
\textbf{Target-acquisition checkpoint comparison.}
For each target paraphrase accuracy, we select the nearest evaluated checkpoint
per seed and method. We report mean
achieved paraphrase accuracy and OOD retention as a percentage of base-model
OOD performance. 
The table abbreviates
QLoRA as QL.
}
\label{tab:matched-acquisition}
\end{table}


\subsection{Model drift is associated with forgetting}

Figure~\ref{fig:drift-diagnostics} shows that configurations with stronger
OOD degradation also exhibit larger drift from the pretrained model. Higher-rank
QLoRA checkpoints have larger symmetric KL divergence and larger effective dense
update magnitudes. The strongest forgetting regime, QLoRA r=64, also has the
largest SVD intruder excess, indicating a larger change in the leading spectral
structure of adapted weight matrices.

These diagnostics are consistent with the behavioral results. Stronger OSM
acquisition is reflected not only in higher paraphrase accuracy but also in larger
distributional and weight-space shifts. High-rank QLoRA
therefore appears to install the target facts through more disruptive updates,
whereas FFT and low-rank QLoRA remain closer to the pretrained model.
This association motivates train-time controls and diagnostics for the trade-off.

\paragraph{Additional math adaptation comparison.}
\begin{table*}[!t]
\centering
\small
\resizebox{\linewidth}{!}{%
    \begin{tabular}{lcccccc|ccc}
    \toprule
    \textbf{Method} & \textbf{MATH-500} & \textbf{Minerva} & \textbf{\begin{tabular}[c]{@{}c@{}}Olympiad\\ Bench\end{tabular}}
    & \textbf{AMC'23} & \textbf{AIME'24} & \textbf{AIME'25}
    & \textbf{\begin{tabular}[c]{@{}c@{}}Math\\ Avg.\end{tabular}}
    & \textbf{\begin{tabular}[c]{@{}c@{}}OOD\\ Avg.\end{tabular}}
    & \textbf{\begin{tabular}[c]{@{}c@{}}OOD\\ Drop $\downarrow$ \end{tabular}} \\
    \midrule
    FFT
    & 78.80 & 34.56 & 44.96 & 60.00 & 20.00 & 16.67
    & 42.50 & 71.12 & 1.71 \\
    QLoRA $r=16$
    & 77.40 & 34.19 & 40.65 & 60.00 & 20.00 & 23.33
    & 42.60 & 70.61 & 2.23 \\
    QLoRA $r=32$
    & 78.60 & 37.13 & 43.92 & 62.50 & 13.33 & 16.67
    & 42.03 & 71.25 & 1.58 \\
    \bottomrule
    \end{tabular}%
}
\caption{\textbf{Math-adaptation results.} Pass@1 scores and averages are in
percent. OOD averages cover HumanEval, IFEval, TruthfulQA, MMLU-Redux, and BBH;
OOD drop is relative to the base Qwen3-4B. 
}
\label{tab:openr1-math-adaptation}
\end{table*}

We run a separate reasoning experiment on a 94k-example subset of
\texttt{OpenR1-Math-220k}~\citep{openr1} to test whether the OSM trend also
appears in a larger skill-adaptation regime. We evaluate Pass@1 on MATH-500
\citep{hendrycks2021measuring}, AIME'24 and AIME'25~\citep{aime}, AMC'23
\citep{amc}, Minerva Math~\citep{lewkowycz2022solving}, and
OlympiadBench~\citep{he-etal-2024-olympiadbench}. OOD degradation uses the same
five-benchmark average as the main experiment, relative to the base Qwen3-4B;
hyperparameters are in Appendix~\ref{app:hyperparam-math}.

Table~\ref{tab:openr1-math-adaptation} shows that the OSM frontier does not
directly transfer to math adaptation. FFT and QLoRA obtain nearly identical
average math performance: 42.50 for FFT, 42.60 for QLoRA $r=16$, and 42.03 for
QLoRA $r=32$. Their OOD drops are also small at 1.71, 2.23, and 1.58 points,
respectively. Math fine-tuning exposes the model to reasoning traces and solution
strategies that may already be supported by pretraining, rather than binding
anonymized entities to novel associations. Consistent with prior task-adaptation
results~\citep{biderman2024lora}, the strong rank-dependent frontier observed on
OSM is not evident in this math setting. In our experiments, it is therefore most
pronounced when adaptation installs new factual associations while preserving
OOD behavior.

\section{Conclusion}
\label{sec:conclusion}
We studied factual acquisition under FFT and QLoRA using an anonymized
OpenStreetMap-derived benchmark. Our results show that QLoRA rank controls an
acquisition--retention trade-off: low-rank adapters preserve general
capabilities but acquire fewer facts, while higher ranks improve same-fact
paraphrase generalization at increasing OOD cost. FFT provides a conservative
baseline, retaining general capabilities well but not reaching the highest
acquisition regime observed with mid-rank QLoRA.
Model-drift diagnostics mirror this pattern: higher-rank QLoRA produces larger
KL divergence, larger effective dense updates, and stronger SVD intruder
effects. 
Thus, PEFT should not be treated as inherently safe for knowledge injection:
adapter rank controls a plasticity trade-off, determining both how much new
factual knowledge is installed and how much pretrained behavior is disturbed.
The unquantized LoRA control suggests that this rank effect does not require
QLoRA quantization, while the much weaker math frontier limits our conclusion to
the present novel-association setting rather than fine-tuning in general.

\section*{Limitations}
\paragraph{Benchmark scope.}
Our OSM dataset comprises 1,938 training examples across 14 small cities, 
so it remains unclear whether the acquisition--retention frontier generalizes 
to larger or more diverse factual corpora. 
Additionally, the use of anonymized synthetic identifiers, 
while useful for controlling pretrained knowledge, 
may not fully reflect real-world knowledge injection scenarios 
where new facts interact with existing world knowledge in richer 
and less controlled ways.
Because some relations have distinct answer types, the benchmark establishes
the acquisition of question-conditioned factual associations but does not fully
separate entity association from abstract relation learning. A stronger test
would use relations with overlapping answer spaces or deliberately conflicting
examples.

\paragraph{Model coverage.}
The main five-seed experiments are conducted with Qwen3-4B, while the
standard-LoRA control uses Qwen3-1.7B.
The shape of the acquisition--retention frontier may differ for larger models, 
models with different pretraining data mixtures, or architectures 
with different weight structures. 
Whether the rank-dependent effects we observe persist at scale remains an open question.

\paragraph{OOD benchmark coverage.}
The five OOD benchmarks used to measure retention
(i.e., HumanEval, IFEval, TruthfulQA, MMLU-Redux, and BBH)
provide a reasonable but not exhaustive proxy for general model capability. 
Retention on other dimensions, such as long-context reasoning or multilingual tasks, 
is not assessed.

\paragraph{Adaptation-method coverage.}
The standard-LoRA control supports the within-adapter rank effect without
quantization, but is limited to one smaller model and ranks 8--32. The main
QLoRA--FFT comparison still differs in quantization and optimization, and the
control lacks matched FFT and QLoRA baselines on Qwen3-1.7B. It therefore does
not isolate every method-level difference.

\paragraph{Math experiment scope.}
The math adaptation comparison is limited to 
two QLoRA ranks ($r \in \{16, 32\}$) and a single epoch of training. 
The conclusion that FFT and QLoRA behave more similarly 
in skill-reinforcement settings therefore rests on a relatively narrow 
hyperparameter sweep, and a fuller rank ablation analogous to 
the OSM experiments would strengthen this claim.

\section*{Ethical Considerations}
The benchmark uses public OpenStreetMap records under the ODbL 1.0 license; full
usage and attribution details are provided in Appendix~\ref{app:osm-license}.
Anonymized task instances remove original entity names and coordinates and
contain no user-level traces. Because the underlying database describes real
places, however, anonymization should not be treated as a guarantee against
geographic re-identification.

\section*{Acknowledgments}
This project was provided with computing HPC and storage resources by GENCI 
at IDRIS thanks to the grant 2025-AD011011668R5 and 2025-AD011017250
on the supercomputer Jean Zay.

\bibliography{custom}

\clearpage

\appendix

\section{Dataset Construction}
\label{sec:dataset}
We construct the dataset from 14 city-level OpenStreetMap (OSM) extracts. The
training split contains 1,938 instruction examples.
We retain points of interest (POIs) and roads with valid, locally unique names,
and derive five atomic relation types: 
POI category, containing city, nearest road, nearest POI, and road-length bucket.

The training data combine direct fact queries, paraphrases of the same facts,
locality-preservation probes, spatial-compositional questions, and inverse city-
signature examples. Spatial examples include four-way nearest-POI selection and
balanced yes/no road-intersection predicates. Examples are sampled with fixed seeds
and relation-balanced quotas to reduce dominance by common POI categories.

For evaluation, we use a held-out paraphrase set of 900 examples constructed from
facts represented in the training data. These evaluation prompts use disjoint lookup,
slot-query, and predicate templates, so they test whether the model recalls the
learned factual associations under different surface forms rather than
memorizing exact training prompts.

Since current LLMs might have some prior knowledge of popular global cities,
we focus on smaller cities with populations between 5,000 and 80,000. To further
reduce the influence of prior knowledge, all names of cities, POIs, and roads are
replaced by synthetic identifiers such as
\texttt{C-TRAIN-001}, \texttt{POI-TRAIN-000001}, and \texttt{ROAD-TRAIN-000001}.
The anonymized task instances contain no source coordinates or user-level data.
Representative examples appear in Appendix~\ref{app:dataset-examples}.

\subsection{Response-format composition}
\label{app:response-formats}
The train and paraphrase splits differ in their proportions of constrained
responses. In particular, yes/no questions make up 6.2\% of the training split
but 13.3\% of the paraphrase split. Treating open-ended exact-match chance as
negligible, four-choice chance as 25\%, and yes/no chance as 50\%, this raises
approximate chance EM from 6.68\% to 10.36\% and partly explains the base-model
difference in Table~\ref{tab:base-prior}.

\begin{table}[t]
\centering
\resizebox{\columnwidth}{!}{%
\begin{tabular}{lrrrr}
\toprule
Split & Open-ended & 4-choice & Yes/no & \begin{tabular}[c]{@{}c@{}}Random\\ Chance \end{tabular}\\
\midrule
Train & \begin{tabular}[c]{@{}c@{}} 1,540 \\ \small(79.5\%)\end{tabular} & \begin{tabular}[c]{@{}c@{}}278 \\ \small(14.3\%)\end{tabular} & \begin{tabular}[c]{@{}c@{}}120 \\ \small(6.2\%)\end{tabular} & 6.68\% \\
Para. & \begin{tabular}[c]{@{}c@{}}647 \\ \small(71.9\%)\end{tabular} & \begin{tabular}[c]{@{}c@{}}133 \\ \small(14.8\%)\end{tabular} & \begin{tabular}[c]{@{}c@{}}120 \\ \small(13.3\%) \end{tabular} & 10.36\% \\
\bottomrule
\end{tabular}%
}
\caption{\textbf{Response-format composition.} Counts and within-split ratios
for training and paraphrase splits, with approximate chance EM for each split.
}
\label{tab:response-formats}
\end{table}

\subsection{OpenStreetMap usage and license}
\label{app:osm-license}
We use OSM database records and geometries---not rendered map tiles---to select
named POIs and roads, determine city membership, compute nearest-neighbor and
intersection relations, and bucket road lengths before anonymization. The source
data are \href{https://www.openstreetmap.org/copyright}{\textcopyright{}
OpenStreetMap contributors}, available under the Open Data Commons Open Database
License (ODbL) 1.0.

\section{Per-benchmark OOD Results at Final Checkpoints}
\label{app:ood-breakdown}

\begin{table*}[t]
\centering
\resizebox{\textwidth}{!}{%
\begin{tabular}{lccccc|c}
\toprule
\textbf{Method} & \textbf{HumanEval} & \textbf{IFEval} & \textbf{TruthfulQA} & \textbf{MMLU-Redux} & \textbf{BBH} & \textbf{OOD Avg.} \\
\midrule
FFT       & 79.2\small{$\pm$2.2} & 80.2\small{$\pm$0.4} & 50.5\small{$\pm$0.6} & 72.7\small{$\pm$0.2} & 70.8\small{$\pm$1.2} & 70.7\small{$\pm$0.7} \\
QLoRA $r=8$  & 78.4\small{$\pm$1.8} & 80.6\small{$\pm$1.3} & 51.6\small{$\pm$1.2} & 73.0\small{$\pm$0.1} & 69.9\small{$\pm$2.1} & 70.7\small{$\pm$0.7} \\
QLoRA $r=16$ & 70.9\small{$\pm$3.0} & 70.3\small{$\pm$1.8} & 50.3\small{$\pm$1.0} & 69.9\small{$\pm$0.2} & 57.7\small{$\pm$4.3} & 63.8\small{$\pm$1.7} \\
QLoRA $r=32$ & 59.9\small{$\pm$4.0} & 34.5\small{$\pm$5.8} & 51.6\small{$\pm$2.0} & 59.0\small{$\pm$4.3} & 34.9\small{$\pm$6.4} & 48.0\small{$\pm$3.6} \\
QLoRA $r=64$ &  23.2\small{$\pm$16.5} & 14.6\small{$\pm$2.4} & 47.7\small{$\pm$1.7} & 37.6\small{$\pm$11.5} & 17.3\small{$\pm$4.0} & 28.1\small{$\pm$6.3} \\
\bottomrule
\end{tabular}%
}
\caption{\textbf{Per-benchmark OOD scores at the final checkpoint (mean $\pm$
standard deviation over five seeds).} Degradation is broad on HumanEval,
IFEval, MMLU-Redux, and BBH;
TruthfulQA is comparatively stable.}
\label{tab:ood-breakdown}
\end{table*}

The final-checkpoint task-level results complement Figure~\ref{fig:acquisition-retention}
and show that the average OOD degradation is not driven by a single benchmark.
HumanEval, IFEval, MMLU-Redux, and BBH decline with increasing QLoRA rank,
whereas TruthfulQA remains comparatively stable.

\section{Details on metrics}
\label{app:metrics}


\paragraph{Symmetric KL.}
Let $p_0(\cdot \mid x_{<t})$ denote the next-token distribution of the
pretrained base model and $p_\theta(\cdot \mid x_{<t})$ the corresponding
distribution of the adapted checkpoint.
We compute
token-level KL divergences under teacher forcing, excluding padding positions.
The reported symmetric KL is
\begin{equation*}
\resizebox{\linewidth}{!}{$
    D_{\mathrm{sym}}(p_0,p_\theta)
    = \tfrac{1}{2}
    \!\left[
    D_{\mathrm{KL}}(p_0 \,\|\, p_\theta) 
    + D_{\mathrm{KL}}(p_\theta \,\|\, p_0)
    \right]
$}
\end{equation*}
averaged over all non-padding tokens and then over batches. 
Instead of using the standard KL that can be dominated by low-probability tokens, 
the symmetric KL emphasizes differences in high-probability regions of the 
distribution, which are more likely to reflect changes in model behavior.
Symmetrization treats
each model in turn as the reference distribution and captures changes in both
directions.
This metric is inspired by \citet{shenfeld2026rls} on distribution shifts. 

\paragraph{Dense RMS drift.}
To compare weight-space drift between FFT and QLoRA, we use the root mean square
(RMS) of the effective dense update, following the scale normalization used in
weight-initialization analyses~\citep{pmlr-v9-glorot10a}. For FFT, the update of a selected
linear module is $\Delta W = W_{\theta} - W_0$. For QLoRA, the effective
merged update is
\[
\Delta W = \frac{\alpha}{r} BA,
\]
where $A$ and $B$ are the LoRA factors, $r$ is the adapter rank, and
$\alpha$ is the LoRA scaling parameter. For a module with $d_{\mathrm{out}}
\times d_{\mathrm{in}}$ dense shape, the module RMS drift is
\[
\mathrm{RMS}(\Delta W)
= \sqrt{
\frac{\|\Delta W\|_F^2}{d_{\mathrm{out}} d_{\mathrm{in}}}
}.
\]
The global dense RMS drift reported in the figures is the same quantity after
summing $\|\Delta W\|_F^2$ and the dense parameter counts over all selected
linear modules:
\[
D_{\mathrm{RMS}}
= \sqrt{
\frac{\sum_m \|\Delta W_m\|_F^2}
{\sum_m d_{\mathrm{out},m} d_{\mathrm{in},m}}
}.
\]

\paragraph{SVD intruder dimensions.}
The SVD diagnostic follows the intruder-dimension construction of
\citet{shuttleworth2024lora}. For each selected linear module, we compute the
top $k$ left singular vectors of the adapted weight matrix and compare each
of them to the top $K$ left singular vectors of the corresponding pretrained
base weight. In our implementation, the defaults are $k=10$ and $K=64$.
For an adapted singular vector $u_i^\theta$, define its best alignment with
the selected base singular vectors as
\[
c_i = \max_{1 \leq j \leq K} |\langle u_i^\theta, u_j^0\rangle|.
\]
For a threshold $\epsilon$, the vector is counted as an intruder when
$c_i < \epsilon$. The diagnostic summary reports the intruder rate,
\[
\mathrm{IntruderRate}_{\epsilon}
=
\frac{\#\{(m,i): c_{m,i}<\epsilon\}}
{\#\{(m,i)\}},
\]
over all selected modules and top adapted singular vectors. 
We use the intruder rate at $\epsilon=0.8$ as the main SVD diagnostic.
To emphasize rank-dependent excess beyond the FFT baseline,
the plotted SVD quantity is
\begin{align*}
\mathrm{IntruderExcess}
&= \mathrm{IntruderRate}^{\mathrm{method}}_{\epsilon=0.8} \\
&- \mathrm{IntruderRate}^{\mathrm{FFT}}_{\epsilon=0.8},
\end{align*}
matched by seed and closest checkpoint step.

\paragraph{Answer log-probability and distractor margin.}
For OSM answer-likelihood diagnostics, we score only the answer continuation
tokens under teacher forcing. Given a prompt \(q\) and answer \(a\), the script
forms the concatenated sequence \([q,a]\), masks out prompt tokens, and reports
the average answer log-probability
\[
\overline{\log p_\theta(a \mid q)}
=
\frac{1}{|a|}
\sum_{t \in a}
\log p_\theta(a_t \mid q, a_{<t}).
\]
The negative log-likelihood is the negative of this average. For the
gold-vs-distractor diagnostic, distractor answers are sampled from examples
in the same split, matching both relation and answer type whenever possible. The reported margin is the
difference between the average log-probability of the gold answer and that of
the sampled distractor; a positive margin means the model assigns higher
teacher-forced likelihood to the gold answer.

\section{Additional dataset examples}
\label{app:dataset-examples}

Below are representative anonymized examples from the training and held-out paraphrase
validation splits.

\paragraph{Training examples.}
\begin{enumerate}
    \item \textbf{Atomic fact.}
    \textit{Question:} In \texttt{C-TRAIN-001}, what type of place is \texttt{POI-
TRAIN-002699}? \\
    \textit{Answer:} \texttt{AMENITY-restaurant}

    \item \textbf{Nearest POI.}
    \textit{Question:} In \texttt{C-TRAIN-001}, which POI is nearest to \texttt{POI-
TRAIN-001802}? \\
    \textit{Answer:} \texttt{POI-TRAIN-001425}

    \item \textbf{Road length bucket.}
    \textit{Question:} In \texttt{C-TRAIN-002}, which length bucket applies to
\texttt{ROAD-TRAIN-027122}? \\
    \textit{Answer:} \texttt{LENGTH-100-200M}

    \item \textbf{Spatial multiple choice.}
    \textit{Question:} In \texttt{C-TRAIN-001}, which POI is closest to 
    \texttt{POI-TRAIN-002343}: \texttt{POI-TRAIN-000318}, \texttt{POI-TRAIN-002124}, 
    \texttt{POI-TRAIN-000864}, \texttt{POI-TRAIN-002699}? \\
    \textit{Answer:} \texttt{POI-TRAIN-002699}

    \item \textbf{Inverse city signature.}
    \textit{Question:} Which city alias matches this local OSM signature?
    \begin{quote}
    \texttt{POI-TRAIN-002699} is a \texttt{AMENITY-restaurant}. \\
    \texttt{POI-TRAIN-000340} is closest to \texttt{POI-TRAIN-000682}. \\
    \texttt{POI-TRAIN-001425} appears in the same city as \texttt{POI-TRAIN-001802}.
    \end{quote}
    \textit{Answer:} \texttt{C-TRAIN-001}
\end{enumerate}

\paragraph{Held-out paraphrase validation examples.}
\begin{enumerate}
    \item \textbf{Slot-style category query.}
    \textit{Question:} Snapshot slot query $\rightarrow$ city: \texttt{C-TRAIN-001};
key: \texttt{POI-TRAIN-001463}; slot: place\_type. \\
    \textit{Answer:} \texttt{AMENITY-school}

    \item \textbf{Nearest-road lookup.}
    \textit{Question:} Map the pair (\texttt{C-TRAIN-002}, \texttt{POI-TRAIN-001715})
to its nearest road. \\
    \textit{Answer:} \texttt{ROAD-TRAIN-019865}

    \item \textbf{Road graph predicate.}
    \textit{Question:} Evaluate this OSM road-graph predicate for city=\texttt{C-
TRAIN-002}: intersects(\texttt{ROAD-TRAIN-030210}, \texttt{ROAD-TRAIN-003453}). Return
yes or no. \\
    \textit{Answer:} \texttt{yes}

    \item \textbf{Paraphrased road-length query.}
    \textit{Question:} Complete this fact: road\_length\_bucket[\texttt{C-TRAIN-002}]
[\texttt{ROAD-TRAIN-017282}] = \\
    \textit{Answer:} \texttt{LENGTH-050-100M}

    \item \textbf{Validation multiple choice.}
    \textit{Question:} OSM relation lookup; city=\texttt{C-TRAIN-009};
relation=nearest\_poi; query=\texttt{POI-TRAIN-001471}; 
choices=[\texttt{POI-TRAIN-000156}, 
\texttt{POI-TRAIN-000138},
\texttt{POI-TRAIN-002766}, \texttt{POI-TRAIN-002758}]. 
Return the matching choice only. \\
    \textit{Answer:} \texttt{POI-TRAIN-000156}
\end{enumerate}

\section{Hyperparameters}
\label{app:hyperparameters}

We report the main hyperparameters for the OSM and math fine-tuning experiments.

\subsection{OpenStreetMap task}
\label{app:hyperparam-osm}
We run a small sweep over $\{2\times10^{-5}, 5\times10^{-5}, 2\times10^{-4}\}$
for QLoRA and $\{2\times10^{-5}, 2\times10^{-4}\}$ for FFT. We select the best
learning rate for each method based on the lowest training loss.

\begin{table}[h]
\centering
\small
\begin{tabular}{lcc}
\toprule
\textbf{Hyperparameter} & \textbf{QLoRA} & \textbf{Full fine-tuning} \\
\midrule
Base model & \multicolumn{2}{c}{Qwen3-4B} \\
Training samples & \multicolumn{2}{c}{1,938} \\
Epochs & \multicolumn{2}{c}{100} \\
Batch size & \multicolumn{2}{c}{16} \\
Learning rate & $2\times10^{-4}$ & $2\times10^{-5}$ \\
LR scheduler & \multicolumn{2}{c}{Linear} \\
Warmup ratio & \multicolumn{2}{c}{0.1} \\
Number of seeds & \multicolumn{2}{c}{5} \\
Optimizer & \texttt{adamw\_8bit} & AdamW \\
LoRA ranks & 8, 16, 32, 64 & -- \\
LoRA alpha & 16, 32, 64, 128 & -- \\
LoRA dropout & 0.05 & -- \\
Target modules & All linear layers & -- \\
\bottomrule
\end{tabular}
\caption{Hyperparameters for the main OpenStreetMap fine-tuning experiments.}
\label{tab:osm-hyperparameters}
\end{table}

\subsection{Math task}
\label{app:hyperparam-math}
We first fine-tune the full model with the same learning rate as in the OSM experiment.
We then run a small sweep over $\{1\times10^{-5}, 2\times10^{-5}\}$ for QLoRA
and select the learning rate with the lowest training loss after one epoch. 

\begin{table}[h]
\centering
\small
\begin{tabular}{lcc}
\toprule
\textbf{Hyperparameter} & \textbf{QLoRA} & \textbf{Full fine-tuning} \\
\midrule
Base model & \multicolumn{2}{c}{Qwen3-4B} \\
Dataset & \multicolumn{2}{c}{\texttt{open-r1-math-220k}} \\
Training samples & \multicolumn{2}{c}{94k} \\
Epochs & \multicolumn{2}{c}{1} \\
Batch size & \multicolumn{2}{c}{$2 \times 16$} \\
Learning rate & $1\times10^{-5}$ & $2\times10^{-5}$ \\
LR scheduler & \multicolumn{2}{c}{Cosine} \\
Optimizer & \texttt{adamw\_8bit} & AdamW \\
LoRA ranks & 16, 32 & -- \\
LoRA alpha & 32, 64 & -- \\
Target modules & All linear layers & -- \\
\bottomrule
\end{tabular}
\caption{Hyperparameters for the additional math adaptation experiments.}
\label{tab:math-hyperparameters}
\end{table}

\end{document}